\documentclass[runningheads]{llncs}
\usepackage[T1]{fontenc}
\usepackage{xcolor} 

\definecolor{modLo}{RGB}{242,246,243}

\definecolor{modHi}{RGB}{63,110,88}

\newcommand{\modmax}{170}      
\newcommand{\modsquaresize}{2.2ex} 

\newcommand{\modsqr}[1]{%
  \begingroup
  \count0=#1\relax
  \count2=\numexpr \count0*85/\modmax\relax 
  \ifnum\count2<0 \count2=0\fi
  \ifnum\count2>100 \count2=100\fi
  \setlength{\fboxsep}{0pt}%
  \colorbox{modHi!\the\count2!modLo}{\phantom{\rule{\modsquaresize}{\modsquaresize}}}%
  \endgroup
}

\usepackage{dblfloatfix}
\usepackage{marvosym}
\usepackage[colorlinks,
            anchorcolor=red,
            citecolor=citecolor, 
            linkcolor=linkcolor,
            ]{hyperref}
\usepackage{amssymb}
\usepackage{amsmath}
\usepackage{graphicx,verbatim}
\usepackage{array}       
\usepackage{multirow}    
\usepackage{booktabs}    
\definecolor{citecolor}{HTML}{0071BC}
\definecolor{linkcolor}{HTML}{ED1C24}
\newcolumntype{P}[1]{>{\centering\arraybackslash}p{#1}}

\newcommand{\msinline}[2]{%
  #1\raisebox{-0.5ex}{$\pm$#2}%
}

\begin{document}
\title{SAGEAgent: A Self-Evolving Agent for Cost-Aware Modality Acquisition in Multimodal Survival Prediction}
\titlerunning{SAGEAgent}
%
\author{Chongyu Qu\inst{1}\and
Can Cui\inst{1}\and
Zhengyi Lu\inst{1}\and
Junchao Zhu\inst{1}\and
Tianyuan Yao\inst{1}\and
Junlin Guo\inst{1}\and
Juming Xiong\inst{1}\and
Yanfan Zhu\inst{1}\and
Yuechen Yang\inst{1}\and
Bennett A. Landman\inst{1,2}\and
Yuankai Huo\inst{1}\Letter}
\authorrunning{C. Qu et al.}
%
\institute{Vanderbilt University, Nashville TN 37235, USA \and
Vanderbilt University Medical Center, Nashville TN, 37232 USA
\email{yuankai.huo@vanderbilt.edu}}

\maketitle              

\begin{abstract}
\textbf{Does every cancer patient truly need a complete diagnostic workup for accurate survival prediction?} In multimodal clinical oncology, diagnostic modalities follow a clinically mandated order of escalating burden---from demographics collected at intake to genomic profiling requiring specialized tissue analysis. Current multimodal survival methods either assume all modalities are available or passively handle missing data, but none actively reason about whether acquiring the next modality is justified for a given patient along this ordered workflow. We formulate this as a sequential decision problem and propose SAGEAgent (\textbf{S}equential \textbf{A}cquisition \textbf{G}uided by \textbf{E}xperience), a self-evolving LLM-based clinical agent that decides which diagnostic modalities to acquire for each patient, balancing predictive accuracy against clinical invasiveness. SAGEAgent reasons about each patient's evolving diagnostic state through clinical tools that translate numerical predictions into text, an episodic memory that retrieves similar past cases, and a semantic memory that accumulates reusable decision patterns from experience. Experiments on a glioma cohort combining TCGA-LGG, TCGA-GBM, and BraTS with four diagnostic modalities demonstrate that SAGEAgent achieves competitive survival prediction accuracy while reducing average acquisition burden by 55\%. Code is publicly available at: \href{https://github.com/Chongyu1117/SAGEAgent}{https://github.com/Chongyu1117/SAGEAgent}
\keywords{Modality Acquisition  \and LLM Agent  \and Survival Prediction.}

\end{abstract}

\section{Introduction}
\label{sec:intro}

\begin{figure}[t]
\includegraphics[width=\textwidth]{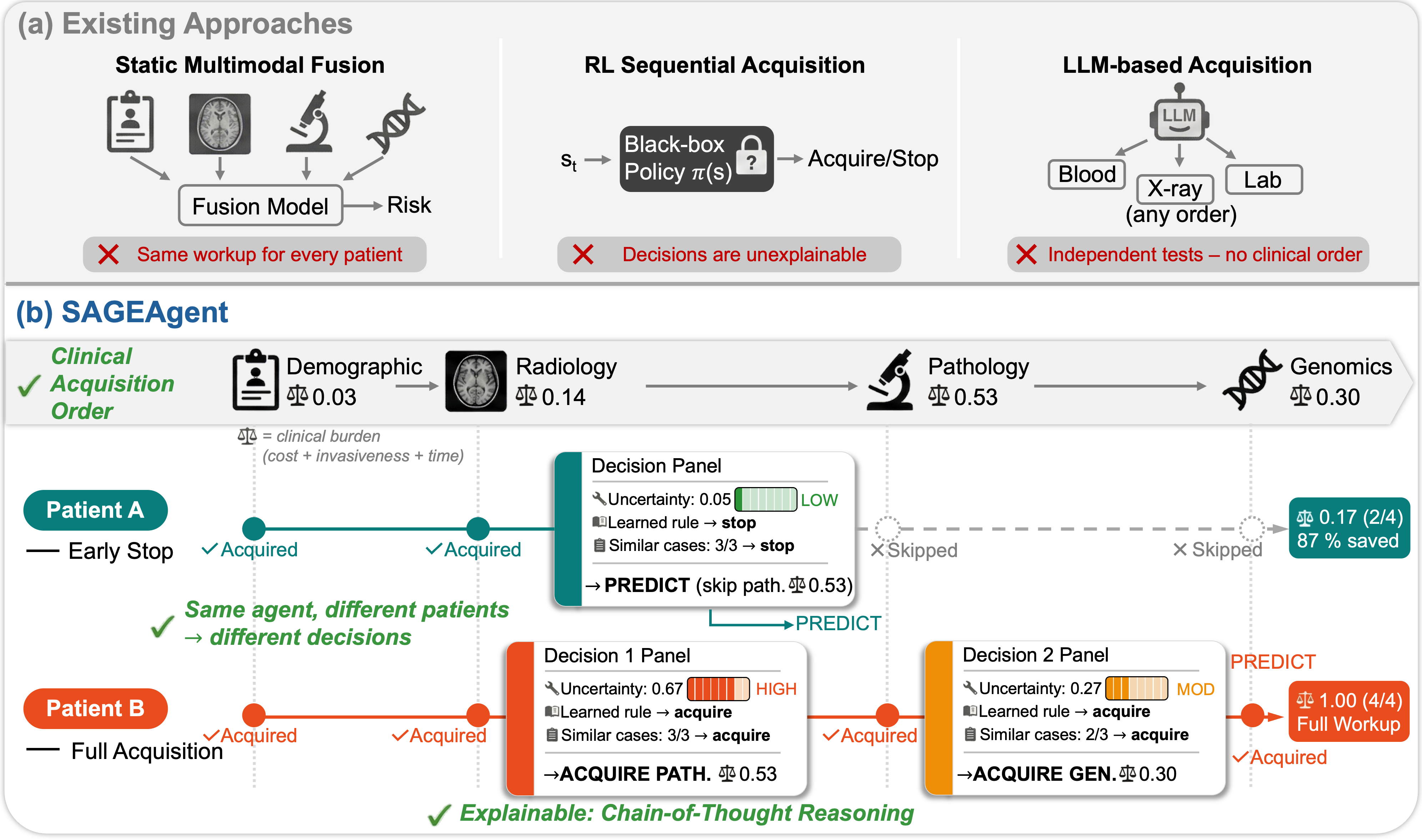}
\caption{\textbf{Overview of modality acquisition strategies.} (a)~Existing approaches: static fusion applies all modalities 
uniformly, RL learns adaptive policies without explanation, and 
LLM-based methods reason over independent tests without clinical 
ordering. (b)~SAGEAgent follows a clinically acquisition 
order, deciding at each stage whether to acquire or stop based on clinical tools, learned rules, and similar past cases. Patient~A 
stops early (83\% burden saved); Patient~B proceeds through all 
modalities.} \label{fig:intro}
\end{figure}

Accurate survival prediction in oncology increasingly relies on integrating multiple diagnostic modalities~\cite{boehm2022harnessing,lipkova2022artificial}, ranging from routine demographics and radiology to invasive biopsy and molecular profiling~\cite{louis20212021}. These modalities follow a strict clinical order: imaging precedes biopsy, which in turn provides the specimen for genomic analysis. Previous multimodal fusion methods~\cite{chen2020pathomic,braman2021deep,cui2022survival} combine all available modalities through a fixed strategy or handle missing data at test time, but treat prediction as a static problem---the model outputs a risk score from whatever is observed, without reasoning about \textbf{whether the next diagnostic step is worth it for this specific patient}.

Sequential test acquisition has been studied from two directions. Reinforcement learning (RL) methods~\cite{bernardino2022reinforcement,saadat2025precise} train cost-aware acquisition policies but produce black-box decisions that limit clinical trust. Large language model (LLM) based approaches~\cite{hager2024evaluation,nori2025sequential,bani2025language} offer transparent reasoning but operate on independent diagnostic tests without accounting for the sequential dependencies inherent in clinical modality acquisition. Neither direction models the clinically mandated ordering (see Fig.~\ref{fig:intro}(a) for a comparison).

We propose SAGEAgent (\textbf{S}equential \textbf{A}cquisition
\textbf{G}uided by \textbf{E}xperience), a training-free, self-evolving
LLM-based framework for cost-aware sequential modality acquisition in
multimodal survival prediction (Fig.~\ref{fig:intro}(b)). At each stage of the clinical order (demographics $\rightarrow$ radiology $\rightarrow$ pathology $\rightarrow$ genomics), SAGEAgent decides whether to acquire or stop using three signal sources: clinical tools that translate numerical predictions into text, an episodic memory that retrieves similar past cases, and a semantic memory that accumulates decision rules through periodic self-reflection---all without gradient updates. 
Evaluated on a glioma cohort combining TCGA-LGG~\cite{pedano2016radiology}, TCGA-GBM~\cite{scarpace2016cancer}, and
BraTS~\cite{menze2014multimodal} with nested 5$\times$5 cross-validation,
SAGEAgent achieves competitive survival prediction accuracy while reducing acquisition burden by 55\% compared to full workup, providing transparent reasoning for each patient's acquisition pathway. 

Our contributions are threefold:
\begin{enumerate}
    \item \textbf{At the learning framework level}, we propose SAGEAgent, a novel framework that formulates cost-aware modality acquisition for multimodal survival prediction as a sequential decision-making problem, while explicitly respecting the clinical ordering of diagnostic procedures.
    \item \textbf{At the training strategy level}, we introduce a dual-memory self-evolution mechanism that autonomously discovers decision rules from experience, eliminating the need for gradient-based updates or hand-crafted priors.
    \item \textbf{At the interpretability level}, we provide fully auditable reasoning pathways in SAGEAgent, where each acquisition decision is grounded in explicit chain-of-thought reasoning, thereby bridging cost-aware modality acquisition with clinical transparency.
\end{enumerate}

\begin{figure*}[t]
\centering
\includegraphics[width=\textwidth]{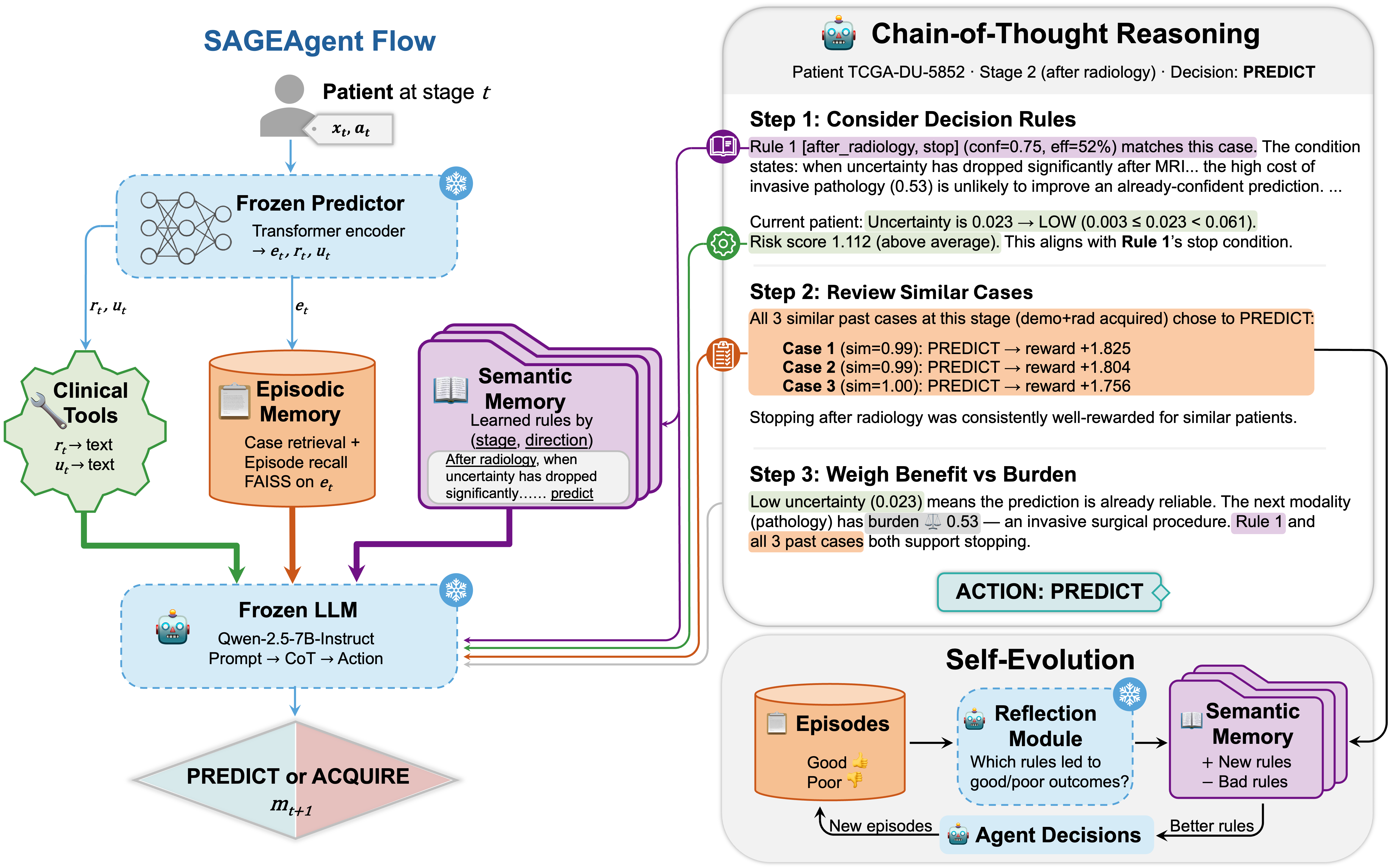}
\caption{\textbf{SAGEAgent architecture and reasoning.}
\textbf{SAGEAgent Flow}: At each stage, the frozen predictor
produces the patient embedding, risk score, and calibrated
uncertainty. Three signal sources feed the frozen LLM: clinical
tools convert numerical outputs to text, episodic memory retrieves
similar patients and past decisions, and semantic memory provides
learned decision rules.
\textbf{Chain-of-Thought Reasoning}: A real reasoning trace after
radiology, with highlights tracing each step to its source
component. The agent synthesizes all three signals and decides to
stop, skipping invasive pathology.
\textbf{Self-Evolution}: Episodes accumulate with outcome labels;
the reflection module analyzes which rules led to good or poor
outcomes and updates semantic memory accordingly, forming a closed
loop without gradient updates.}
\label{fig:method}
\end{figure*}

\section{Method}
\label{sec:method}

\subsection{Problem Formulation}

Consider a patient with $N$ diagnostic modalities
$\{m_1, \ldots, m_N\}$ arranged in a clinically mandated acquisition
order $m_1 \rightarrow m_2 \rightarrow \cdots \rightarrow m_N$. For
glioma, this order is demographics $\rightarrow$ radiology $\rightarrow$
pathology $\rightarrow$ genomics, reflecting real clinical
dependencies~\cite{louis20212021}. Each modality $m_i$ carries a clinical
burden $b(m_i) \in [0, 1]$, quantified via multi-criteria decision
analysis (MCDA)~\cite{guitouni1998tentative} over monetary cost, turnaround time, patient
invasiveness, and infrastructure requirements (details in
Sec.~\ref{sec:setup}). At decision stage $t \in \{1, \ldots, N{-}1\}$,
the agent has already acquired the first $t$ modalities. The decision is
binary: acquire~$m_{t+1}$ (incurring burden $b(m_{t+1})$), or stop and
predict. The goal is to maximize prediction accuracy while minimizing
total acquisition burden $B_t = \sum_{i=1}^{t} b(m_i)$.
Fig.~\ref{fig:method} provides an overview of the proposed framework.

\subsection{Multimodal Predictor with Calibrated
Uncertainty}\label{sec:predictor}

\noindent \textbf{Survival predictor.}
The prediction backbone is a Transformer-based multimodal encoder that
handles missing modalities via learnable mask
tokens~\cite{devlin2019bert,cui2022survival}. It maps each patient's
available modalities to a fused embedding $\mathbf{e}_t \in
\mathbb{R}^d$, from which a Cox proportional hazards
head~\cite{cox1972regression} produces a scalar risk score $r_t$. The
predictor is trained with modality dropout and \textbf{frozen during agent operation}. Architecture and
training details are in Sec.~\ref{sec:impl}.

\noindent \textbf{Calibrated uncertainty head.}
On top of the frozen predictor, we train a lightweight uncertainty head
that estimates whether the current prediction would change if more
modalities were acquired. Given the embedding $\mathbf{e}_t$ and
acquisition mask $\mathbf{a}_t$, it outputs a calibrated probability:
\begin{equation}
u_t = \sigma\!\bigl(f_\phi([\mathbf{e}_t,\, \mathbf{a}_t])\bigr)
\label{eq:uncertainty}
\end{equation}
where $f_\phi$ is a lightweight MLP and $\sigma$ is the sigmoid
function. The training label is 1 when the absolute difference between
the current and full-modality risk scores exceeds a threshold $\tau$,
and 0 otherwise. Training data is generated by enumerating
clinical-order prefix masks on complete-modality patients. The resulting
$u_t \in [0,1]$ is patient-specific and mask-specific, giving the agent
a calibrated signal of whether additional data is likely to matter.

\subsection{SAGEAgent: LLM-Based Acquisition Agent}\label{sec:agent}

SAGEAgent is built on a frozen LLM that receives no gradient updates.
At each stage, it gathers diagnostic signals, retrieves relevant
experience, and outputs a binary decision (acquire or predict) with a
chain-of-thought explanation. Its behavior improves solely through two
memory systems following the episodic--semantic distinction in cognitive
science~\cite{tulving1972episodic}.

\noindent \textbf{Clinical decision tools.}
Two tools translate the frozen predictor's numerical outputs into text.
The \textit{Uncertainty Tool} reports $u_t$ from
Eq.~\ref{eq:uncertainty} with a categorical level (very low to very
high) based on quintile thresholds from the training distribution. The
\textit{Survival Predictor Tool} reports the risk score $r_t$ and its
position relative to the training cohort (below/around/above average). Both tools present outputs as natural-language sentences so
that all reasoning occurs in text space.

\noindent \textbf{Episodic memory.}
The episodic memory provides case-based reasoning through two
complementary retrieval mechanisms.
\textit{Case retrieval} draws from a FAISS-indexed~\cite{douze2025faiss} database
of training patients, returning the $k$ nearest neighbors by cosine
similarity on $\ell_2$-normalized embeddings together with their event status
(survived or deceased) and survival time. This provides a static
clinical reference: which training patients resemble the current one,
and how long did they survive.
\textit{Episode recall} draws from the agent's own
past decisions. When the agent acquires $m_{t+1}$, it receives a stage
reward $R_{\text{stage}} = -b(m_{t+1}) + \max(0,\, u_t - u_{t+1})$ that
penalizes cost and rewards uncertainty reduction. When the agent stops at
stage $t$, a terminal reward $R_{\text{term}} = (1 - u_t) - B_t$ favors
early stopping at low uncertainty. Retrieval is stage-matched and
re-ranked by the sum of cosine similarity
$\text{sim}(\mathbf{e}_t, \mathbf{e}_j)$ and min-max normalized episode
reward $\tilde{r}_j \in [0,1]$, returning the top-$k$ episodes, including the action taken and its outcome.

\noindent \textbf{Semantic memory and self-evolution.}
The semantic memory stores interpretable decision rules discovered
through periodic self-reflection. Each rule is indexed by acquisition
stage $t$ and action direction $d \in \{\text{stop}, \text{acquire}\}$.
After each episode, we record whether the agent followed or violated
each relevant rule. Every $N_r$ episodes (one reflection cycle),
episodes are ranked by total reward, and those in the top and bottom
quartiles are labeled good and poor respectively. Rule effectiveness is
the fraction of followed episodes that led to good outcomes. The
reflection module receives the effectiveness report and proposes new
rules, which are verified for direction clarity and novelty. Rules that
consistently lead to poor outcomes are deprecated, forming a closed loop
where rules are discovered, validated, and either reinforced or removed.

\section{Experiments \& Results}
\label{sec:experiments}

\subsection{Dataset and Evaluation Protocol}
\label{sec:setup}

\noindent \textbf{Dataset and split.}
We evaluate on a glioma cohort of 962 patients
(803 TCGA-LGG/GBM~\cite{pedano2016radiology,scarpace2016cancer} and 159
BraTS~\cite{menze2014multimodal}) spanning four diagnostic modalities in
clinical order: demographics (Demo.), radiology (Rad.), pathology
(Path.), and genomics (Gen.), each encoded into a 32-dimensional
feature vector by pretrained extractors~\cite{cui2022survival}.
Of the 962 patients, 170 possess all four modalities and form the
evaluation set. We adopt nested $5{\times}5$ cross-validation.
The outer loop splits 170 patients into 5 folds (136/34) for agent
training and evaluation. Each inner 5-fold trains the predictor on
the 136 patients plus 792 partial-modality patients as augmentation
via modality dropout, yielding 25 independent pipelines.
Per-patient decisions are aggregated by majority vote
($\geq$3/5 inner models). We report mean~$\pm$~std across 5
outer folds.

\noindent \textbf{Clinical burden.}
We quantify modality cost via MCDA~\cite{guitouni1998tentative} over
monetary cost, turnaround time, invasiveness, and infrastructure
requirements. After normalization, per-modality burdens are 0.03
(demographics), 0.14 (radiology), 0.53 (pathology), and 0.30
(genomics), totaling $B{=}1.0$ for a full workup.

\noindent \textbf{Metrics.}
We report C-index~\cite{harrell1982evaluating}, average burden, and
hypervolume $\text{HV} = (c - 0.5) \times (1.0 - b)$ which
captures the accuracy--cost trade-off as a single scalar.

\begin{table*}[t]
\centering
\caption{\textbf{Core results on the glioma survival cohort
(170 complete-modality patients).}
Top~(a)~C-index in ascending order. RL-based methods form a lower
cluster around 0.73, while the remaining six methods achieve
C-index above 0.8.
Top~(b)~C-index (solid) and burden (hatched) for the six methods
with C-index\,$>$\,0.8.
Bottom: per-method breakdown.  \modsqr{170} = 170 patients, \modsqr{5} = 5 patients, with shades varying linearly between these values.
$\dagger$\,=\,predictor backbone used by SAGEAgent.}
\label{tab:results}
\setlength{\tabcolsep}{3pt}
\includegraphics[width=\textwidth]{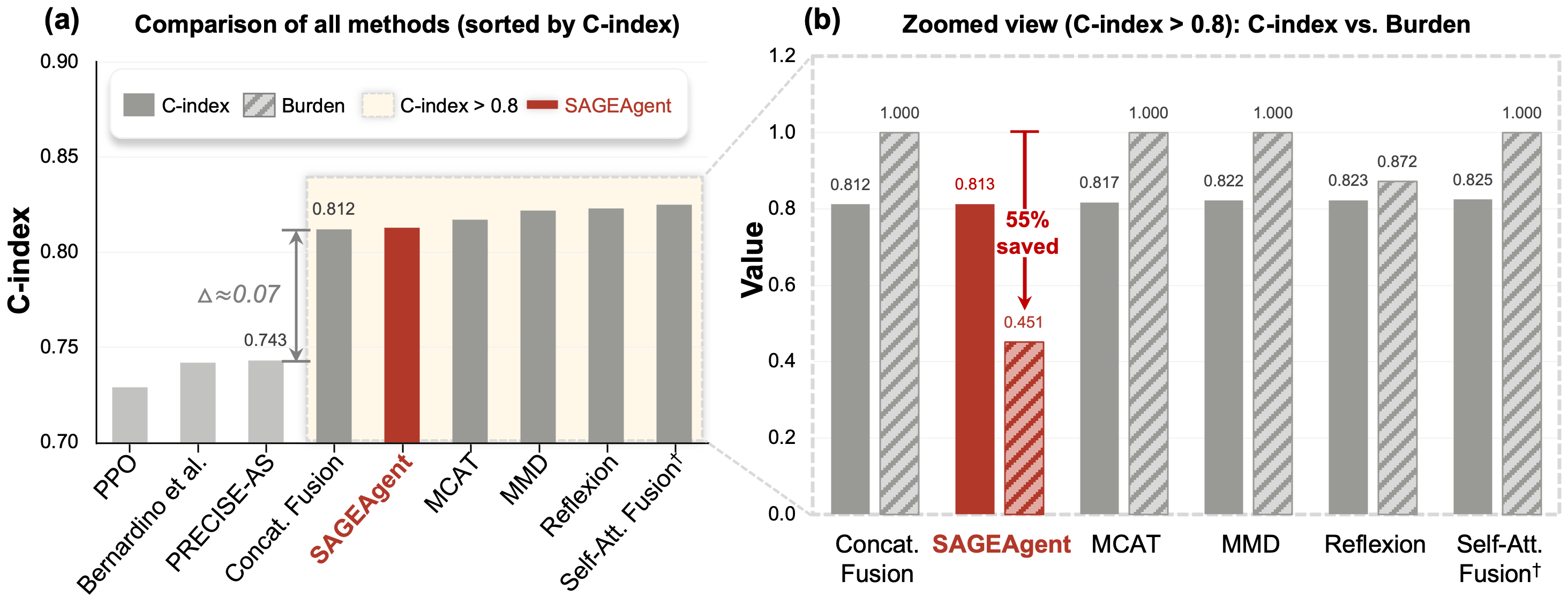}
\begin{tabular}{p{0.14\linewidth} p{0.23\linewidth} p{0.13\linewidth} P{0.14\linewidth} P{0.062\linewidth} P{0.05\linewidth} P{0.05\linewidth} P{0.05\linewidth}}
\toprule
Category & Method & C-index$\uparrow$ & Burden$\downarrow$ (\%saved) & Demo. & Rad. & Path. & Gen. \\
\midrule
Static   & Concat. Fusion
           & \msinline{0.812}{.040} & 1.000 (0\%) & \modsqr{170} & \modsqr{170} & \modsqr{170} & \modsqr{170} \\
Fusion   & MCAT~\cite{chen2021multimodal}
           & \msinline{0.817}{.046} & 1.000 (0\%) & \modsqr{170} & \modsqr{170} & \modsqr{170} & \modsqr{170} \\
         & MMD~\cite{cui2022survival}
           & \msinline{0.822}{.032} & 1.000 (0\%) & \modsqr{170} & \modsqr{170} & \modsqr{170} & \modsqr{170} \\
         & Self-Att. Fusion$^\dagger$
           & \msinline{0.825}{.050} & 1.000 (0\%) & \modsqr{170} & \modsqr{170} & \modsqr{170} & \modsqr{170} \\
\midrule
RL-based       & Bernardino et al.~\cite{bernardino2022reinforcement}
           & \msinline{0.742}{.058} & 0.125 (88\%) & \modsqr{170} & \modsqr{60} & \modsqr{14} & \modsqr{2} \\
Acquisition & PRECISE-AS~\cite{saadat2025precise}
           & \msinline{0.743}{.059} & 0.127 (87\%) & \modsqr{170} & \modsqr{61} & \modsqr{14} & \modsqr{2} \\
         & PPO~\cite{schulman2017proximal}
           & \msinline{0.729}{.070} & 0.124 (88\%) & \modsqr{170} & \modsqr{49} & \modsqr{14} & \modsqr{5} \\
\midrule
LLM-based      & Reflexion~\cite{shinn2023reflexion}
           & \msinline{0.823}{.049} & 0.872 (13\%) & \modsqr{170} & \modsqr{169} & \modsqr{165} & \modsqr{107} \\
Acquisition & \textbf{SAGEAgent}
           & \msinline{0.813}{.046} & 0.451 (55\%) & \modsqr{170} & \modsqr{166} & \modsqr{85} & \modsqr{11} \\
\bottomrule
\end{tabular}
\end{table*}
\subsection{Implementation Details}
\label{sec:impl}

\noindent \textbf{Predictor.}
The backbone is a 2-layer Transformer encoder ($d{=}128$, 4 heads)
with per-modality MLP projections from $\mathbb{R}^{32}$ to
$\mathbb{R}^{128}$ and learnable mask tokens for missing modalities,
trained with Cox partial likelihood, reconstruction, and alignment
losses under 50\% modality dropout (${\sim}$458K parameters). The
calibrated uncertainty head (Eq.~\ref{eq:uncertainty}) adds
${\sim}$6.5K parameters trained with BCE loss, with $\tau{=}0.3$
set by calibration quality (AUROC\,=\,0.908, ECE\,=\,0.110) on
inner-fold validation after post-hoc temperature
scaling~\cite{guo2017calibration}.

\noindent \textbf{SAGEAgent} uses Qwen-2.5-7B-Instruct~\cite{yang2025qwen3} with fully
frozen weights, running on a single NVIDIA A6000 GPU. During
experience accumulation, each training patient is processed for 3
episodes, yielding ${\sim}$408 episodes per fold. Episodic memory
retrieves $k{=}3$ stage-matched neighbors via FAISS~\cite{douze2025faiss}.
Semantic memory maintains up to 10 active rules with reflection
triggered every 10 patients.

\subsection{Core Results}
\label{sec:results}

Table~\ref{tab:results} and compare SAGEAgent
against static fusion baselines and acquisition methods under the
nested 5$\times$5 CV protocol described in Sec.~\ref{sec:setup}.

\noindent \textbf{RL methods reduce cost at the expense of accuracy.}
A C-index gap of approximately 0.07 separates the nine methods into
two clusters.(Table~\ref{tab:results} top~(a)) All three RL baselines reduce burden to
${\sim}$0.12 but drop C-index to ${\sim}$0.73, effectively
discarding informative modalities. Static fusion methods require
full workup by design, and Reflexion reduces burden by only 13\%.

\noindent \textbf{SAGEAgent maintains accuracy while reducing burden
by 55\%.}
As shown in Table~\ref{tab:results} top~(b), SAGEAgent achieves a
C-index of 0.813, within 0.012 of the full-modality backbone, while
reducing burden to 0.451. Compared to Reflexion, SAGEAgent achieves
$4.2{\times}$ greater burden reduction with a C-index difference of
only 0.010. Among compared methods, SAGEAgent is the only one that
achieves meaningful burden reduction without compromising accuracy.

\begin{table*}[t]
\centering
\caption{\textbf{Component ablation.} Each row adds one component to
the Base LLM. All configurations use frozen Qwen-2.5-7B-Instruct
with majority-vote aggregation. Bold indicates best result. \modsqr{170} = 170 patients, \modsqr{5} = 5 patients, with shades varying linearly.}
\label{tab:ablation}
\setlength{\tabcolsep}{2.5pt}
\begin{tabular}{p{0.26\linewidth} P{0.14\linewidth} P{0.16\linewidth} P{0.07\linewidth} P{0.06\linewidth} P{0.06\linewidth} P{0.06\linewidth} P{0.06\linewidth}}
\toprule
Configuration & C-index$\uparrow$ & Burden$\downarrow$ (\%saved) & HV$\uparrow$ & Demo. & Rad. & Path. & Gen. \\
\midrule
Base LLM
  & \msinline{0.827}{.040} & 0.798 (20\%) & 0.066 & \modsqr{170} & \modsqr{170} & \modsqr{168} & \modsqr{59} \\
\quad + Tools
  & \msinline{0.830}{.035} & 0.843 (16\%) & 0.052 & \modsqr{170} & \modsqr{170} & \modsqr{165} & \modsqr{90} \\
\quad + Tools + Episodic
  & \msinline{0.814}{.027} & 0.611 (39\%) & 0.122 & \modsqr{170} & \modsqr{170} & \modsqr{117} & \modsqr{43} \\
\quad + Tools + Semantic
  & \msinline{0.819}{.036} & 0.706 (29\%) & 0.094 & \modsqr{170} & \modsqr{170} & \modsqr{139} & \modsqr{58} \\
\textbf{SAGEAgent}
  & \msinline{0.813}{.046} & \textbf{0.451} (55\%) & \textbf{0.172} & \modsqr{170} & \modsqr{166} & \modsqr{85} & \modsqr{11} \\
\bottomrule
\end{tabular}
\end{table*}
\subsection{Ablation Study}
\label{sec:ablation}

Table~\ref{tab:ablation} isolates the contribution of each
component by progressively adding tools, episodic memory, and
semantic memory to the frozen LLM. Adding tools alone raises burden
from 0.798 to 0.843 because without past experience to contextualize
uncertainty and risk signals, the LLM defaults to acquiring more.
Episodic memory provides the largest individual burden reduction
(0.843$\to$0.611) by surfacing cases where early stopping succeeded,
while semantic memory offers a complementary effect
(0.843$\to$0.706) through interpretable stopping criteria. Combined,
SAGEAgent achieves the lowest burden (0.451) and highest HV (0.172),
a $2.6{\times}$ improvement over the Base LLM.

\begin{table*}[t]
\centering
\caption{\textbf{Rules discovered by semantic memory across five folds.} Eff.\,=\,fraction of rule-following episodes
with top-quartile reward (random baseline\,=\,0.25). $n$\,=\,total episodes where the rule was followed across folds.
Stability indicates whether the rule, once discovered, was ever deprecated and re-proposed during training.}
\label{tab:evolution}
\setlength{\tabcolsep}{3pt}
\begin{tabular}{
  p{0.6\linewidth}
  P{0.05\linewidth}
  P{0.05\linewidth}
  P{0.09\linewidth}
  P{0.09\linewidth}}
\toprule
Discovered Rule & Folds & $n$ & Eff. & Stability \\
\midrule
\multicolumn{5}{l}{\textit{Stopping rules}} \\[2pt]
\cmidrule(lr){1-5}
``After radiology, predict now \ldots risk error is small
  and pathology cost outweighs potential gain.''
  & 5/5 & 876 & .46--.52 & Stable \\[4pt]
  \cmidrule(lr){1-1}
``After pathology, predict now \ldots prediction quality
  is high and genomics cost is not justified.''
  & 5/5 & 387 & .39--.62 & Stable \\
\midrule
\multicolumn{5}{l}{\textit{Acquisition rules}} \\[2pt]
\cmidrule(lr){1-5}
``After demographics, acquire radiology \ldots uncertainty
  is high and imaging cost is low.''
  & 4/5 & 493 & .38--.67 & Stable \\[4pt]
  \cmidrule(lr){1-1}
``After radiology, acquire pathology \ldots uncertainty
  remains high and risk error is large.''
  & 4/5 & 353 & .44--.58 & Churned \\
\bottomrule
\end{tabular}
\end{table*}

\subsection{Self-Evolution Analysis}
\label{sec:evolution}

The ablation in Sec.~\ref{sec:ablation} shows that semantic memory
reduces burden, but does not reveal what rules it learns. We now
examine whether the discovered rules are consistent across folds.
Each outer fold runs 10 reflection cycles from an empty rule set
with no shared initialization.

\noindent \textbf{Semantic memory discovers consistent rules across
folds.}
Table~\ref{tab:evolution} lists the four rules emerging in at
least four of five folds, all achieving effectiveness well above
the 0.25 random baseline and collectively covering over 2{,}100
decisions. Each fold generates 10.8 candidate rules on average, of
which 4.6 survive pruning.

\noindent \textbf{Stopping rules are easier to learn than acquisition
rules.}
The two stopping rules emerge in the first two reflection cycles and
are never deprecated. In contrast, the acquisition rule ``after
radiology, acquire pathology'' accumulates 13 deprecated instances
across folds before a stable version survives in four of five.
Reflection logs reveals a two-phase pattern:
cycles 1--2 establish core stopping rules that persist unchanged,
while cycles 3--10 are dominated by acquire-rule exploration where
most proposals are deprecated due to near-zero effectiveness.

\section{Conclusion}
\label{sec:conclusion}
We presented SAGEAgent, a training-free LLM-based framework for
cost-aware sequential modality acquisition in multimodal survival
prediction. On a glioma cohort, SAGEAgent reduces acquisition burden
by 55\% while maintaining accuracy within 0.012 C-index of
full-modality fusion, with each decision grounded in auditable
reasoning. Notably, the agent learns to skip invasive pathology for
half of the cohort, suggesting that a substantial fraction of glioma
patients may not require biopsy for reliable prognostication.
Self-evolution analysis reveals that stopping rules converge
reliably while acquisition rules are inherently harder to learn.
Future work includes evaluation on larger multi-site cohorts, other cancer types, and prospective clinical validation.

%
%
\bibliographystyle{splncs04}
\bibliography{Paper-0802}

\end{document}